\documentclass{article}

    \PassOptionsToPackage{numbers, compress}{natbib}
    \usepackage[final]{neurips_2020}
\usepackage[utf8]{inputenc} % allow utf-8 input
\usepackage[T1]{fontenc}    % use 8-bit T1 fonts
\usepackage{hyperref}       % hyperlinks
\usepackage{url}            % simple URL typesetting
\usepackage{booktabs}       % professional-quality tables
\usepackage{amsfonts}       % blackboard math symbols
\usepackage{nicefrac}       % compact symbols for 1/2, etc.
\usepackage{microtype}      % microtypography
\usepackage{amsmath}

\usepackage{algorithm}
\usepackage{booktabs} % for professional tables
\usepackage{subcaption}
\usepackage{caption}
\usepackage{xcolor}
\usepackage[noend]{algpseudocode}
\usepackage{wrapfig}
\usepackage{adjustbox}
\usepackage{xcolor}

\definecolor{forestgreen}{rgb}{0.14, 0.55, 0.14}

\renewcommand{\paragraph}[1]{\vspace{0.20ex}\noindent\textbf{#1}}

\def\eg{\emph{e.g.~}}

\def\ie{\emph{i.e.~}}

\title{Sensitivity of Average Precision to Bounding Box Perturbations}
\author{Ali Borji \\
Quintic AI, San Francisco, CA \\
\texttt{aliborji@gmail.com} }

\begin{document}

\maketitle

\begin{abstract}

Object detection is a fundamental vision task. It has been highly researched in academia and has been widely adopted in industry. Average Precision (AP) is the standard score for evaluating object detectors. Our understanding of the subtleties of this score, however, is limited. 
% It could well be that performance saturation of models on COCO dataset due to shortcomings of the mAP score. 
Here, we quantify the sensitivity of AP to bounding box perturbations and show that AP is very sensitive to small translations. Only one pixel shift is enough to drop the mAP of a model by 8.4\%. The mAP drop over small objects with only one pixel shift is 23.1\%. The corresponding numbers when ground-truth (GT) boxes are used as predictions are 23\% and 41.7\%, respectively. 
These results explain why achieving higher mAP becomes increasingly harder as models get better. 
We also investigate the effect of box scaling on AP. Code and data is available at 
\url{https://github.com/aliborji/AP_Box_Perturbation}.

\end{abstract}

\section{Motivation}

Tremendous success has been achieved in the area of object detection. Object detection models have come a long way but performance is still low compared to other vision tasks such as object recognition\footnote{The best score in terms of mean AP (mAP) on COCO validation set is about 63.2\%. Please see \url{https://paperswithcode.com/sota/object-detection-on-coco-minival}.}. The reasons are still unknown~\cite{borji2020empirical,borji2019empirical,bolya2020tide}. 
We suspect that it may be partly due to how models are evaluated, and more specifically, the way AP is computed. AP is a complicated score and it is difficult to get all details right in implementing this score~\cite{padilla2020survey}. Compared to huge efforts spent on building models, less attention has been paid to understanding the scores for object detection. Here, we study how errors in predicted boxes impact the accuracy. 

% understanding the shortcomings of models or designing new ones. 

\begin{figure}[htbp]
    \centering
    \includegraphics[width=.48\textwidth]{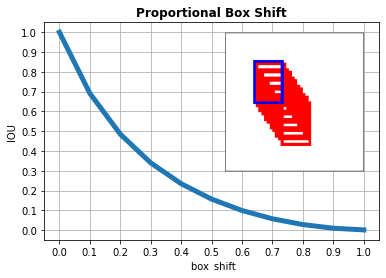}
    \includegraphics[width=.48\textwidth]{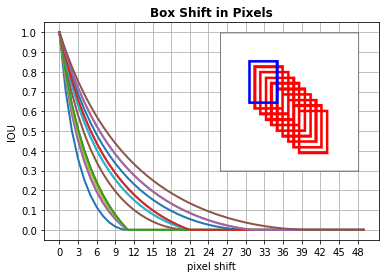}    

    \caption{Sensitivity of IOU to bounding box perturbation. Left) Proportional shift, Right) Fixed pixel shift (here for two pixel shift).}
    \label{fig:iou}
\end{figure}

% It is not clear whether AP has started to saturate, whether a small improvement in AP (\eg 56.1 vs. 56 mAP) is meaningful

% Average precision (AP), for instance, is a popular metric for evaluating the accuracy of object detectors by estimating the area under the curve (AUC) of the precision × recall relationship. Object detection is an extensively studied topic in the field of computer vision. Different approaches have been employed to solve the growing need for accurate object detection models [1]. In most competitions, the average precision (AP) and its derivations are the metrics adopted to assess the detections and thus rank the teams.

\begin{figure}[t]
    \centering
    \includegraphics[width=.48\textwidth]{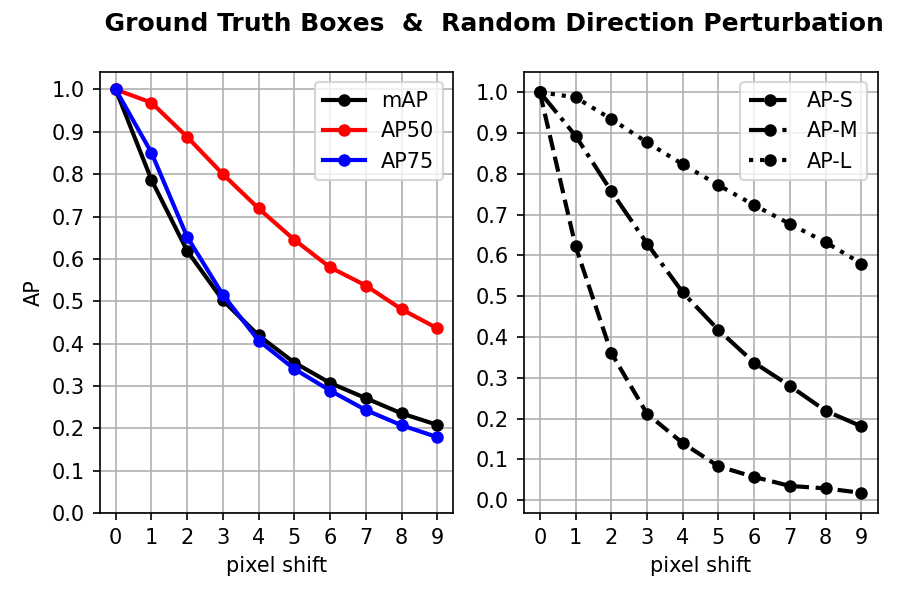}
    \includegraphics[width=.48\textwidth]{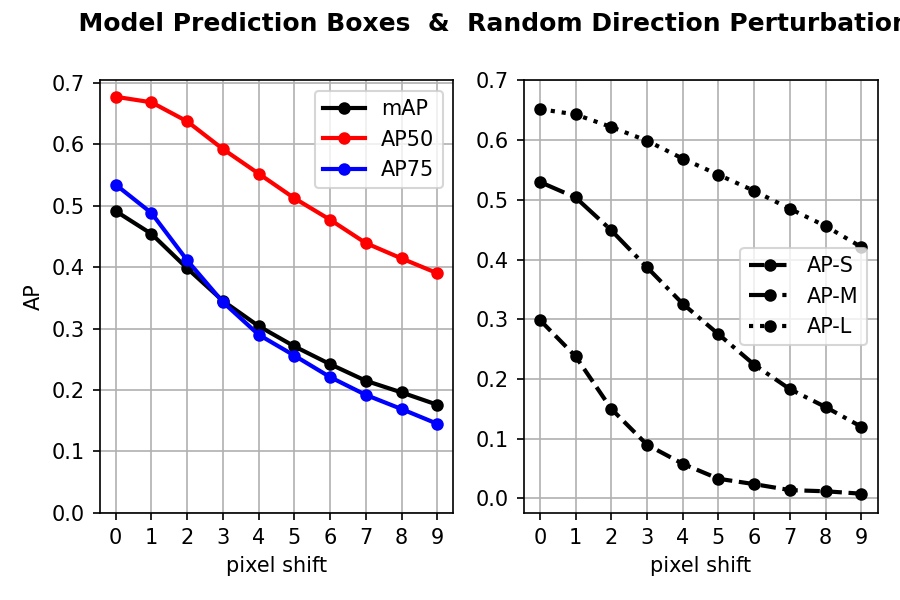}    

    \caption{Sensitivity of AP to bounding box perturbation.}
    \label{fig:randomDirection}
\end{figure}

\section{Experiments and Results}
We first study how shifting the bounding box impacts the IOU. Following from that, we then study the effect of box translation on AP.

\subsection{Sensitivity of IOU to bounding box translation}

We randomly generated a number of bounding boxes with different widths and heights. We then shifted the bounding boxes diagonally in two ways. In the first approach, a box is shifted in steps proportional to the box size. Let the tuple $(x_l, y_l, x_r, y_r)$ represent the original bounding box A. The shifted box B is then computed as follows:

\begin{center}
B = ($x_l$ + offset $\times$ W, $y_l$ + offset $\times$ H, $x_r$ + offset $\times$ W, $y_r$ + offset $\times$ H])
\end{center}

where W and H are width and height of the box, respectively. The offset varies in the range $[0,1]$ in steps of 0.1\footnote{numpy.linspace(0,1,11)}. 

In the second approach, we simply add a fixed offset to the box coordinates as follows:
\begin{center}
B = ($x_l$ + offset, $y_l$ + offset, $x_r$ + offset, $y_r$ + offset) 
\end{center}
Here, offset varies for a number of pixels (\eg 10).

The IOU of the boxes A and B over random boxes is shown in Fig.~\ref{fig:iou} for both approaches. The higher the box shift, the lower the IOU. The drop in IOU is not linear. The slope is sharper for small shifts than larger ones. As will be shown later this also applies to the AP. The profile of decline in the first approach is the same for all boxes regardless of the bounding box shape since it is proportional to width and height. It does, however, depend on the bounding box shape in the second approach (the right panel in Fig.~\ref{fig:iou}). As expected, small boxes are impacted much more than the larger ones for the same amount of pixel shift.

\subsection{Sensitivity of AP to bounding box translation}
Here, we only consider box shifts in pixels since it is more intuitive and is easier to interpret (\ie approach two above). Eight directions for shifting the box \texttt{\{left, right, top, down, top left, top right, bottom left, bottom right\}} are included. The magnitude of shift (offset) varies in pixels from 0 to 10. Two regimes are considered. In the first regime, called random direction perturbation, a bounding box is randomly shifted in one of the eight directions. In the second regime, all boxes are shifted in the same direction. 

Over the MS COCO validation set~\cite{lin2014microsoft}, we consider the ground truth bounding boxes as predictions and give them score 1. We then compute the AP for these predictions in both regimes. 
Results for the first regime are shown in Fig.~\ref{fig:randomDirection}. As expected, increasing the shift lowers the performance across all three APs: mAP, AP50, and AP75. The AP50 is less hindered compared to the other two APs since a small shift does not change the IOU50 much. We observe that:

\begin{itemize}
\item {\bf 1 pixel shift of GT box (GT used as prediction) in a random direction lowers the mAP about 23.8\%, and 2 pixels shift lowers it about 41.6\%, }
\item 1 pixel shift of GT box in a random direction lowers the AP50 about 3.8\%, and 2 pixels shift lowers it about 13.1\%,
\item 1 pixel shift of GT box in a random direction lowers the AP75 about 18.7\%, and 2 pixels shift lowers it about 30.6\%,
\item {\bf 1 pixel shift of GT box in a random direction lowers the mAP over large, medium, and small objects 1.7\%, 13.4\%, and 41.7\%, respectively.}
\end{itemize}

With predictions of the MaskRCNN~\cite{he2017mask} (with their original scores), we find that:

\begin{figure}[t]
    \centering
    \includegraphics[width=.48\textwidth]{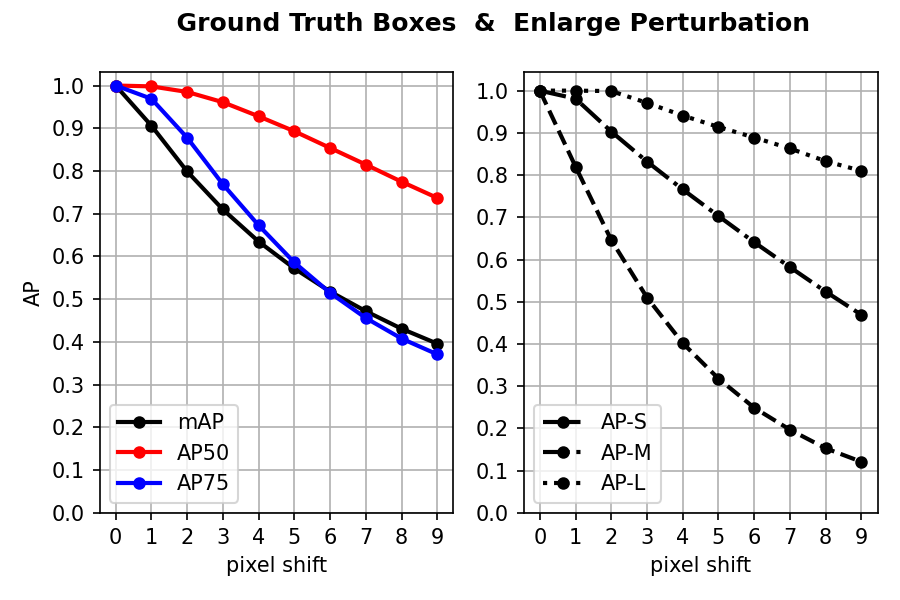}
    \includegraphics[width=.48\textwidth]{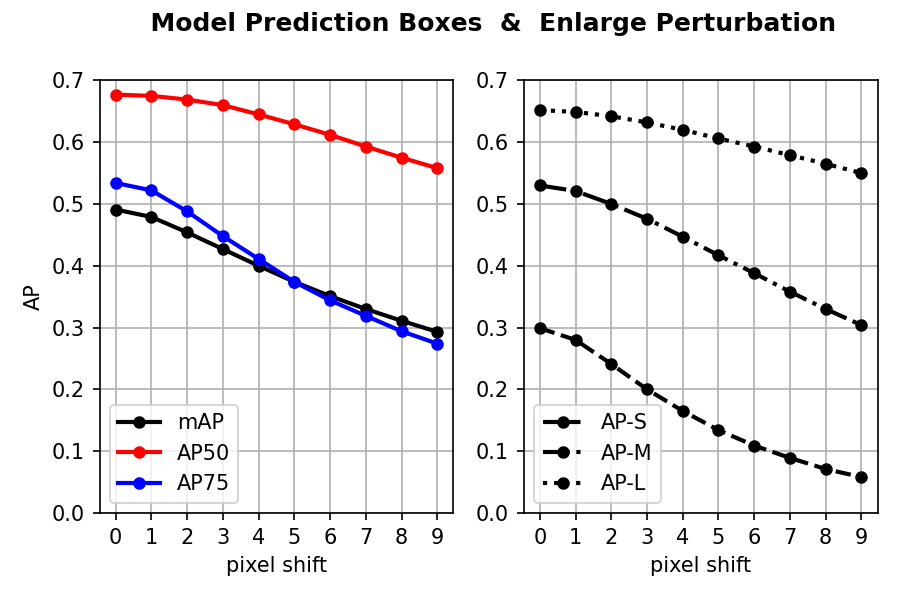} \\

    \includegraphics[width=.48\textwidth]{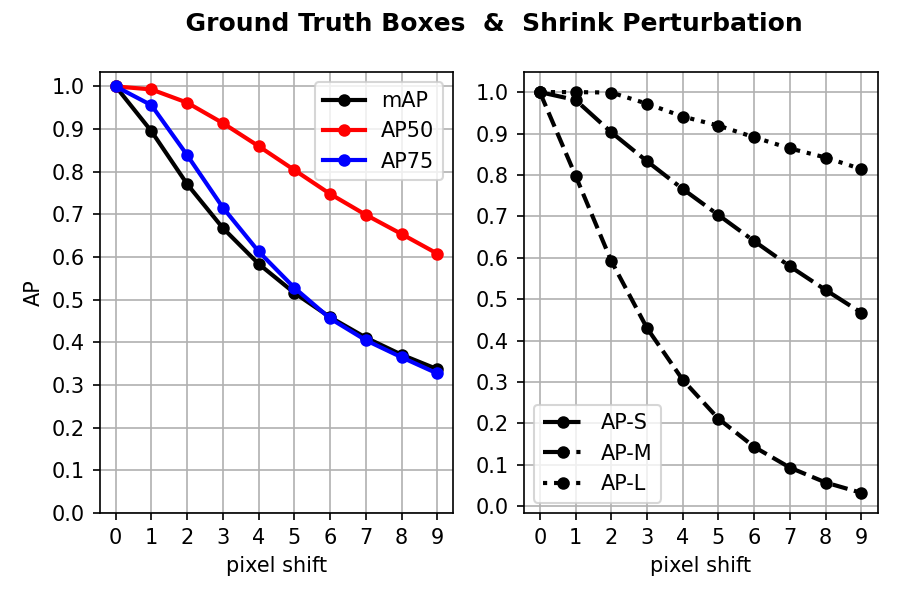} 
    \includegraphics[width=.48\textwidth]{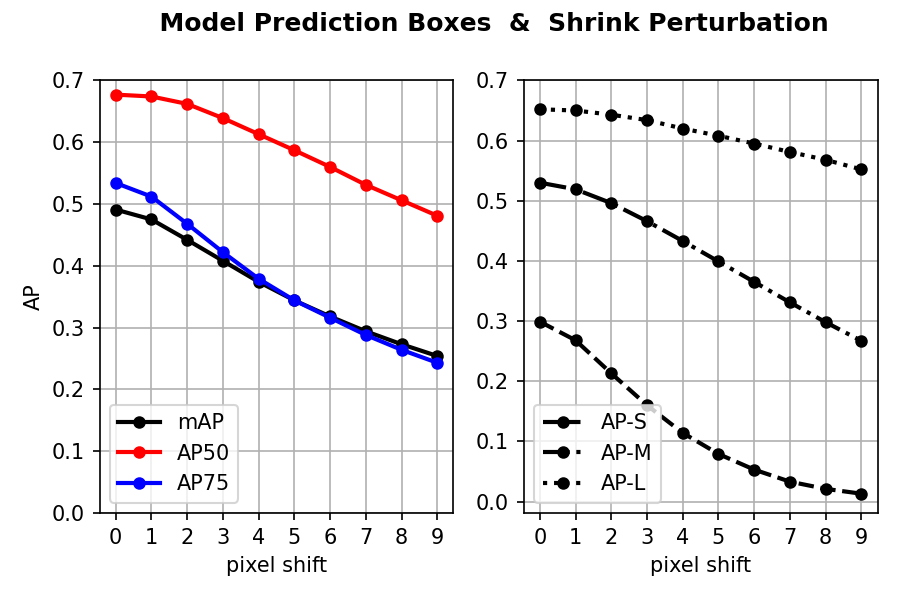}

    \caption{Sensitivity of AP to bounding box scaling.}
    \label{fig:scale}
\end{figure}

\begin{itemize}
\item {\bf 1 pixel shift of model prediction in a random direction lowers the mAP about 8.4\%, and 2 pixels shift lowers it about 20.1\%,}
\item 1 pixel shift of model prediction in a random direction lowers the AP50 about 1.5\%, and 2 pixels shift lowers it about 7.1\%, 
\item 1 pixel shift of model prediction in a random direction lowers the AP75 about 9.6\%, and 2 pixels shift lowers it about 25.4\%, 
\item {\bf 1 pixel shift of model prediction in a random direction lowers the mAP over large, medium, and small objects 1.38\%, 5.85\%, and 23.1\%, respectively.}
\end{itemize}

The threshold for an object to be considered small in COCO is area $32^2$. Just one pixel shift in the box is enough to drop the mAP more than 40\% using GT boxes. Results for when boxes are all shifted in the same direction is shown in Appendix for all directions. As expected diagonal directions hinder AP more than horizontal or vertical directions. Drop in AP seems to be symmetric (\eg same amount for left and right directions).

% The mAP is hindered severely over small objects.  

% small objects: area < 32^2
% medium objects: 32^2 < area < 96^2
% large objects: area > 96^2

\subsection{Sensitivity to bounding box scaling}
We also investigate two other transformations. In the first one, called enlarging, box A is scaled according to the following formula:

\begin{center}
B = ($x_l$, $y_l$, $x_r$ + offset, $y_r$ + offset) 
\end{center}
Here, top left corner of the box is held fixed and the right corner is shifted outwards. In the second type of scaling, called shrinking, the bounding box is shrunk as follows:

\begin{center}
B = ($x_l$, $y_l$, $x_r$ - offset, $y_r$ - offset) 
\end{center}

As above, the offset varies from 0 to 10 pixels. Results are shown in Fig.~\ref{fig:scale}. It seems like both box transformations have similar effects. One pixel box enlarging or shrinking lowers the mAP about 10\% using GT bounding boxes (20\% over small objects) and 4\% using model predictions (8\% over small objects). Scaling, the way we do here, hinders the AP less than box translations mainly because the box is changed less.

\section{Discussion and Conclusion}

We find that AP is overly sensitive to bounding box precision. This may explain why performance in object detection is saturating. Small errors in predicted boxes are much more costly in higher APs than lower APs. For example, improving from 0.9 to 0.91 mAP is much harder than improving from 0.4 to 0.41. 

Our investigation suggests that we should seek other complementary ways to evaluate object detectors. Some possibilities include a) designing alternative scores (as in~\cite{oksuz2018localization}), b) asking humans to judge how good a bounding box is, and c) using detected boxes in downstream tasks (\eg object classification).

The analyses of this sort can also give insights into the complexity of the datasets (\ie a measure of image or dataset clutter). For examples, datasets for which AP drops quickly with box shifting may suggest that objects are occluding each other a lot. Same analyses can also be done for 3D object detection and other tasks such as object segmentation.

% 1) Using human evaluation on subset of data
% 2) 

% Here, we only considered 

% if objects occlude each other a lot, then small perturbations may change AP a lot. (i.e. a measure of clutter in a DB).

% {\bf Future work.}

% This entails that perhaps we should also seek other ways to evaluate detectors (\eg other scores such as xx, human evaluation, or downstream tasks). 

\bibliographystyle{plain}
\bibliography{refs}

\clearpage
\appendix

\section{Sensitivity of AP to bounding box perturbation}

\begin{figure}[htbp]
    \centering
    \includegraphics[width=.48\textwidth]{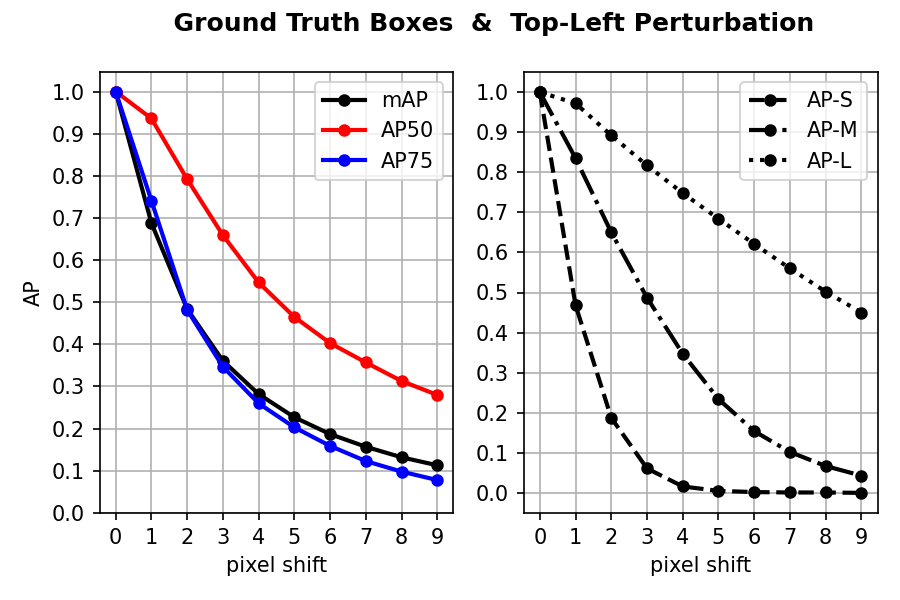}
    \includegraphics[width=.48\textwidth]{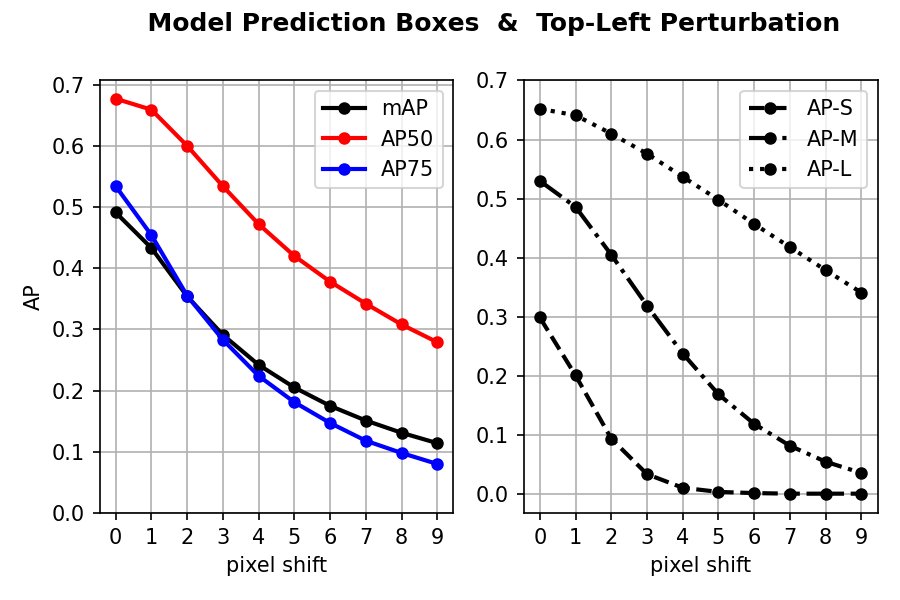} \\ 
    
    \includegraphics[width=.48\textwidth]{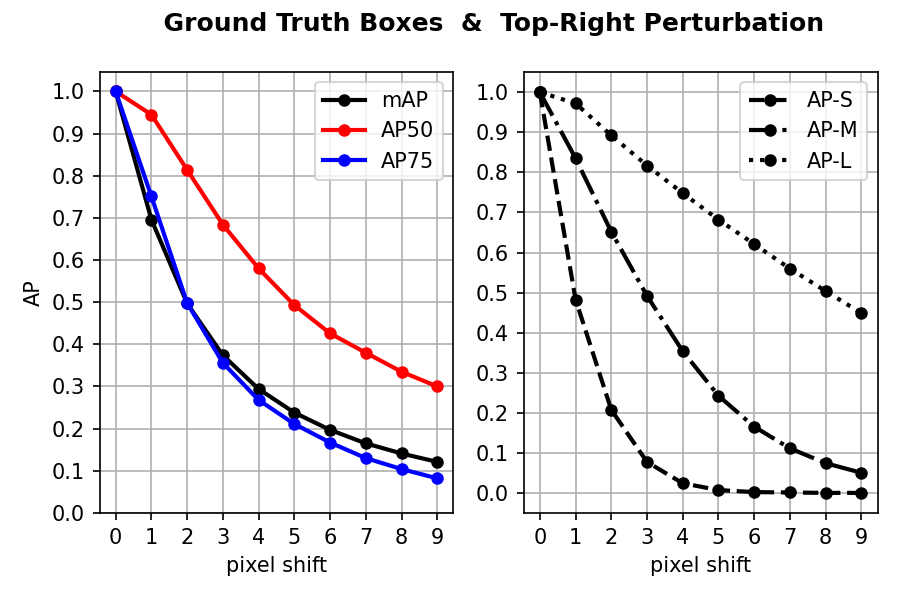} 
    \includegraphics[width=.48\textwidth]{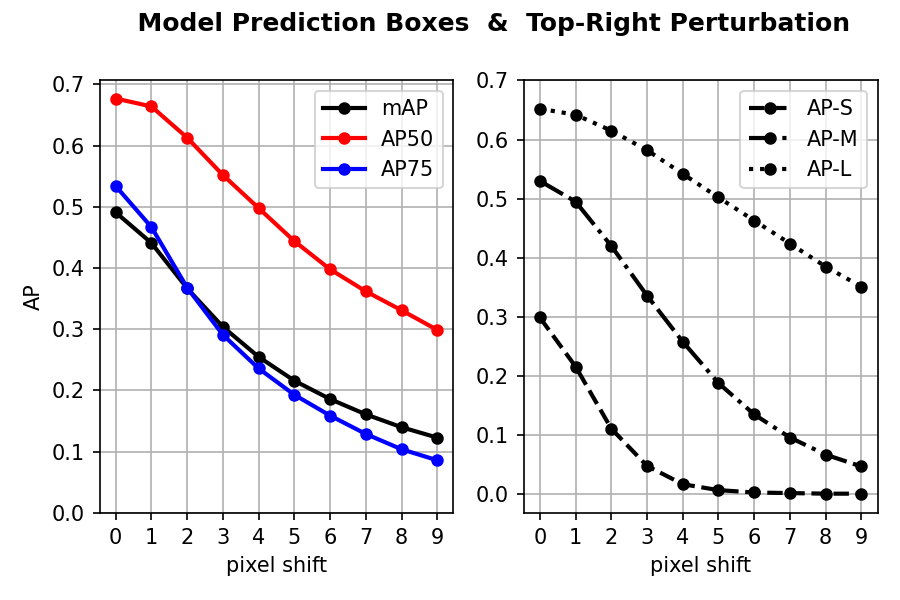} \\
    
    \includegraphics[width=.48\textwidth]{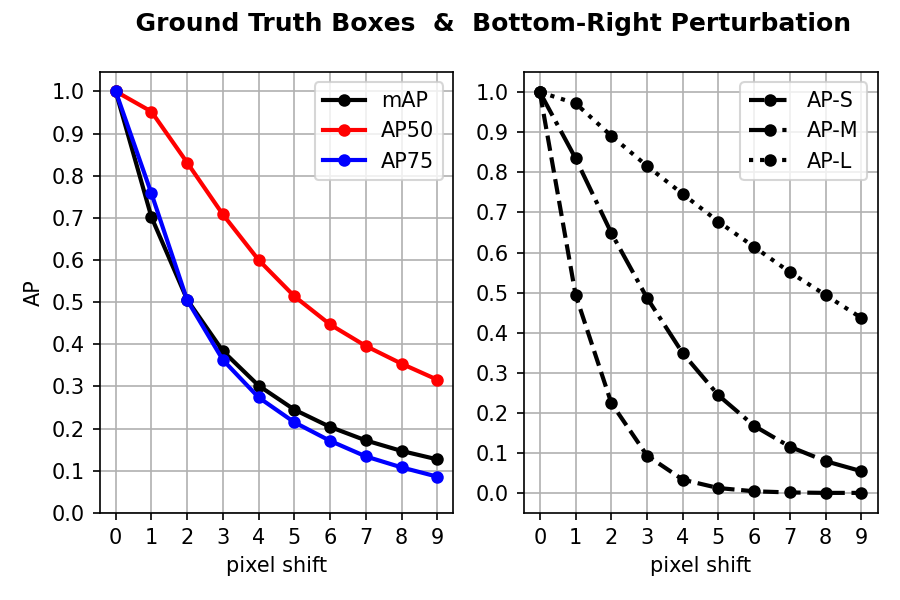} 
    \includegraphics[width=.48\textwidth]{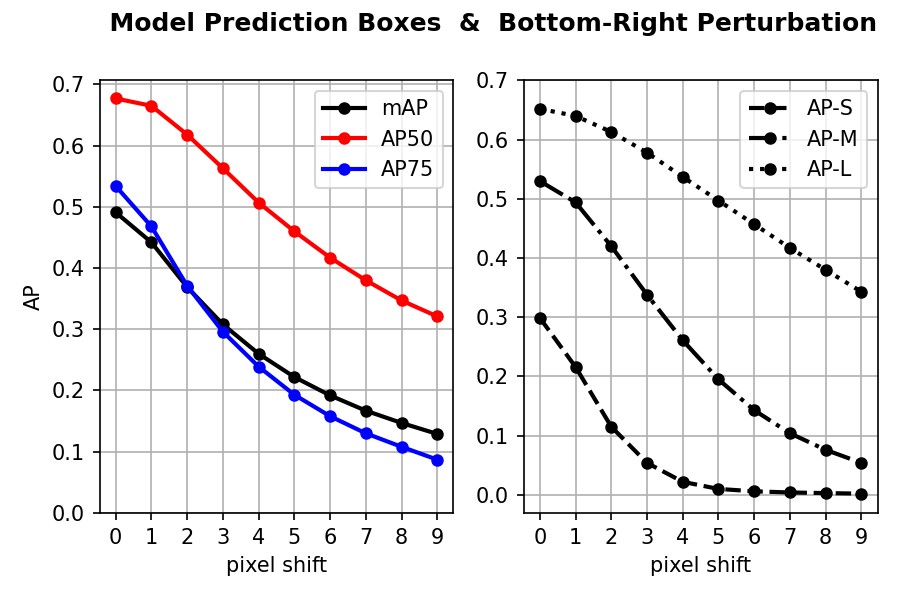} \\    
    
    \includegraphics[width=.48\textwidth]{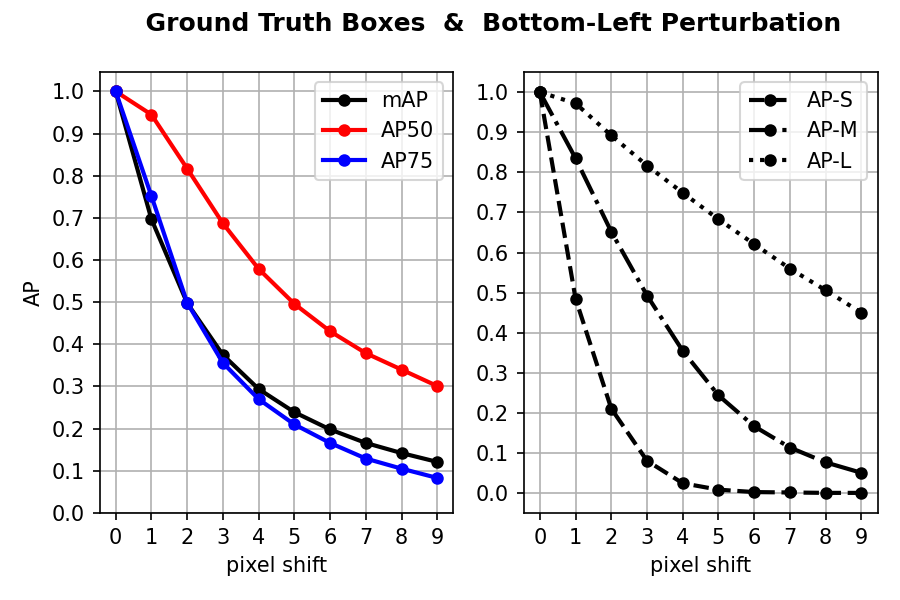} 
    \includegraphics[width=.48\textwidth]{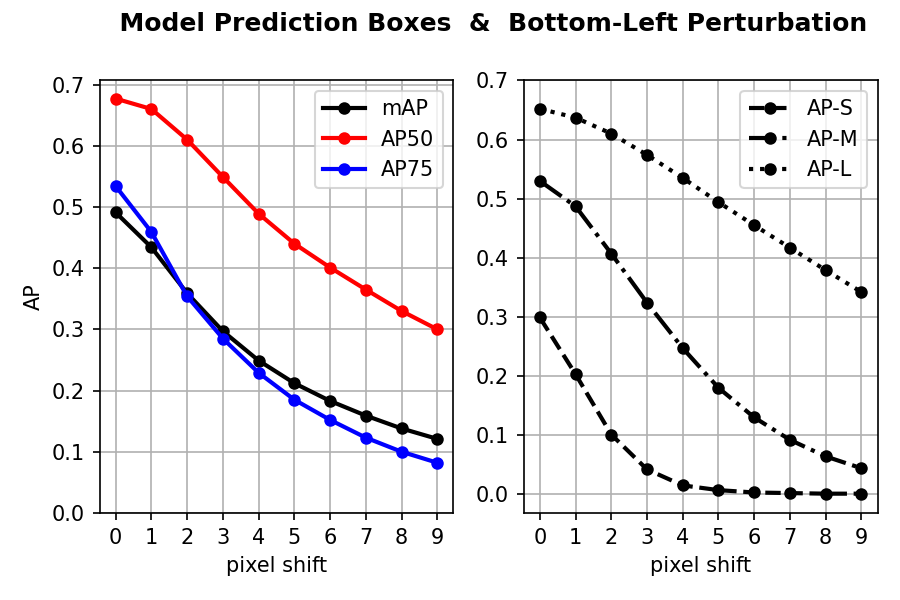} \\
    
    \caption{Sensitivity of AP to bounding box perturbation (left: Ground Truth boxes as predictions, right: MaskRCNN predictions).}
    \label{fig:GT}
\end{figure}

\begin{figure}[htbp]
    \centering
    
    \includegraphics[width=.48\textwidth]{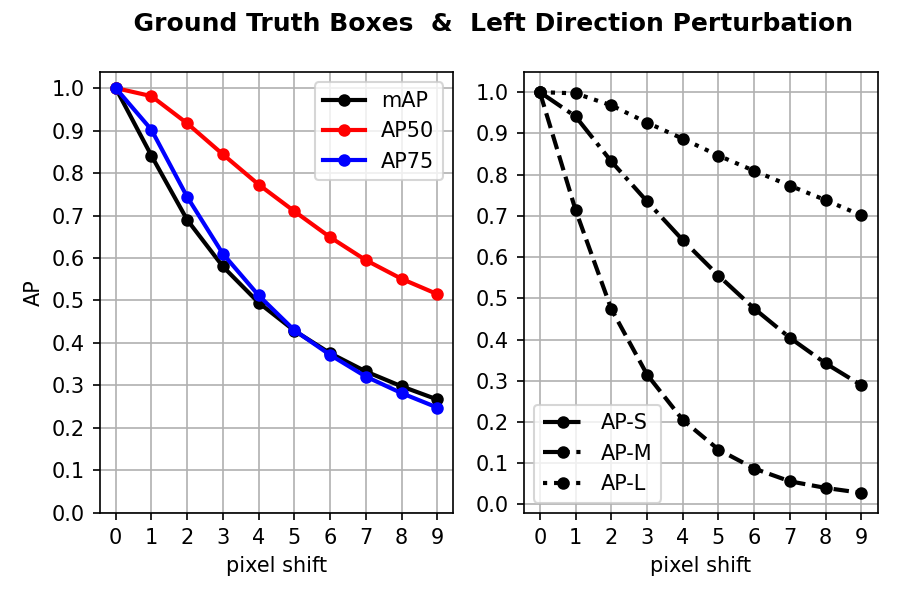}
    \includegraphics[width=.48\textwidth]{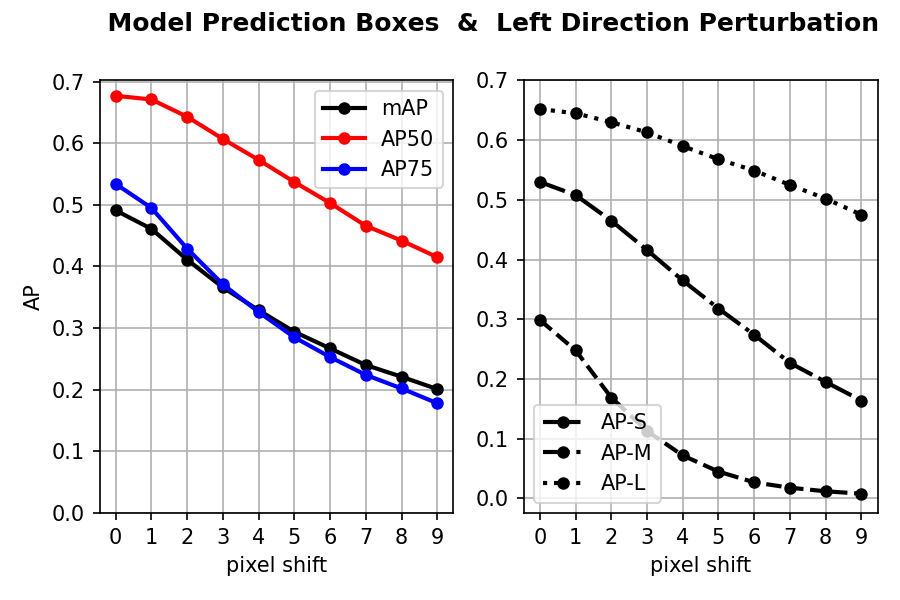}    \\
    
    \includegraphics[width=.48\textwidth]{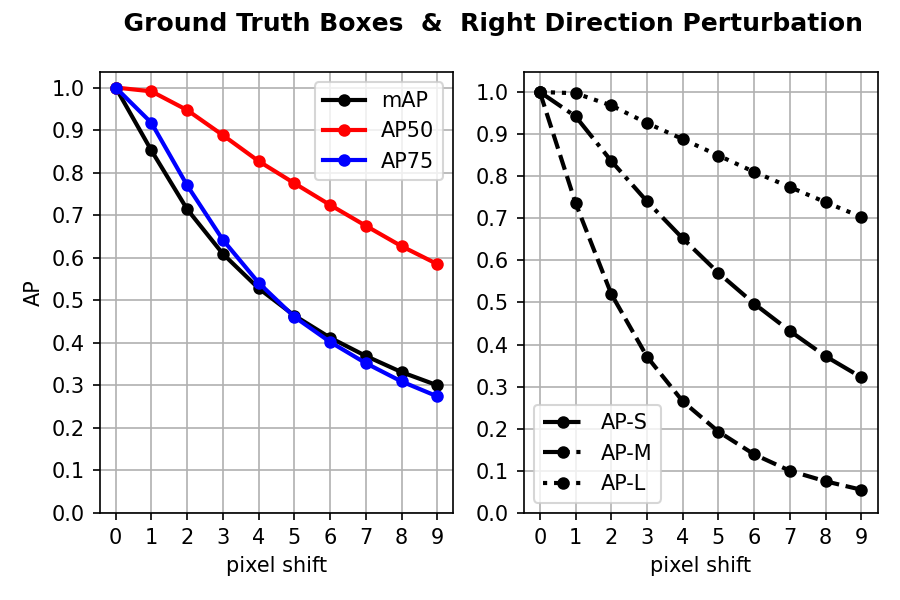} 
    \includegraphics[width=.48\textwidth]{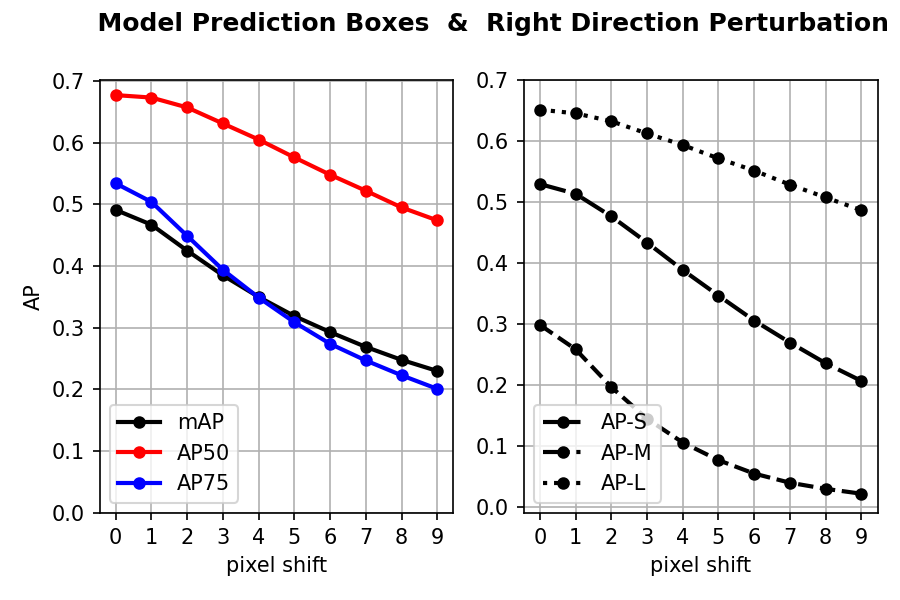} \\    
    
    \includegraphics[width=.48\textwidth]{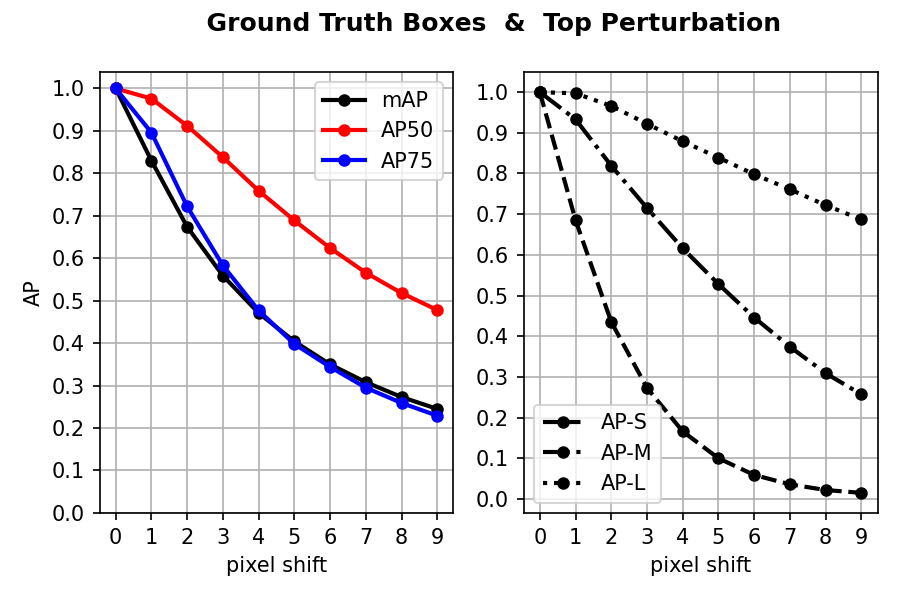} 
    \includegraphics[width=.48\textwidth]{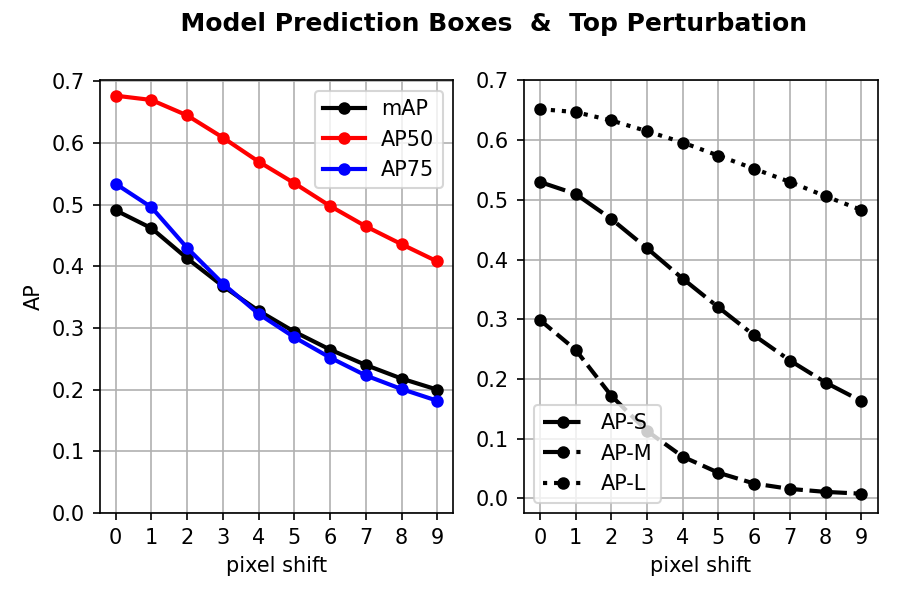} \\
    
    \includegraphics[width=.48\textwidth]{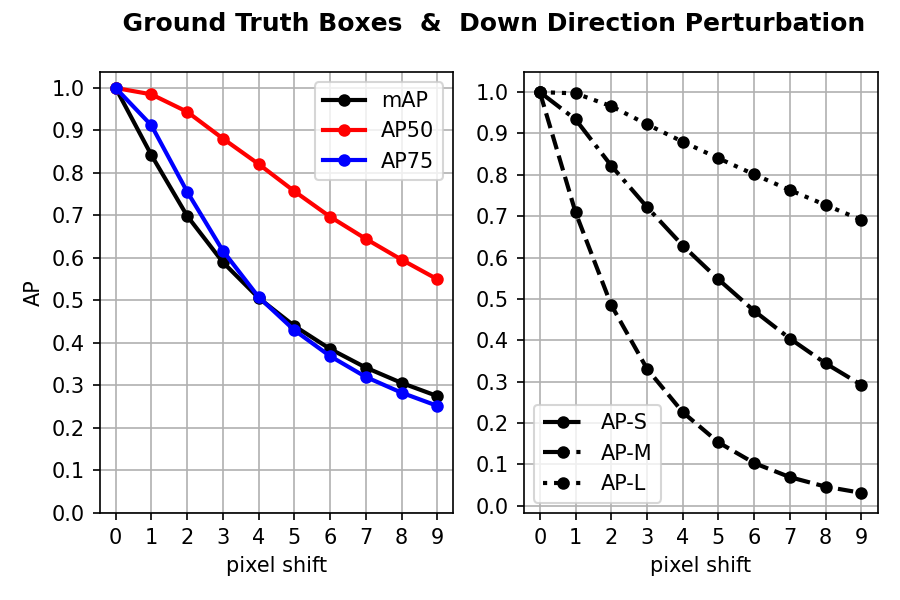} 
    \includegraphics[width=.48\textwidth]{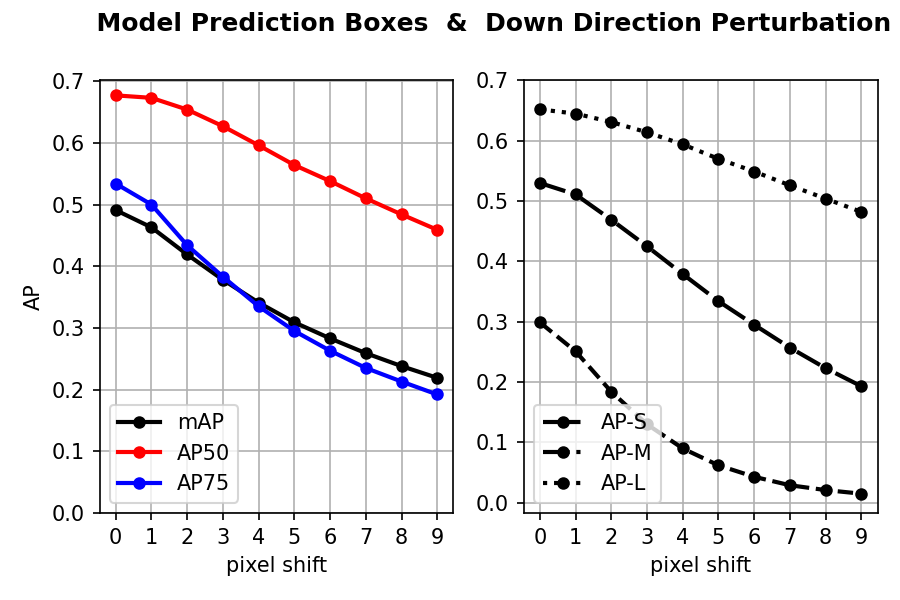}

    \caption{Sensitivity of AP to bounding box perturbation (continued).}
    \label{fig:Model}

\end{figure}

\end{document}